%% file: arxiv.tex
\documentclass[10pt,twocolumn,letterpaper]{article}

\usepackage[pagenumbers]{wacv} % To force page numbers, e.g. for an arXiv version

\input{preamble}

\usepackage{booktabs}
\usepackage{multirow}
\usepackage{graphicx} % for resizebox
\usepackage{amssymb} % for \checkmark
\usepackage{algorithm}
\usepackage{algpseudocode}
\usepackage{listings}
\usepackage[table]{xcolor}
\usepackage{pifont}
\newcommand{\xmark}{\ding{55}}

\definecolor{wacvblue}{rgb}{0.21,0.49,0.74}
\usepackage[pagebackref,breaklinks,colorlinks,allcolors=wacvblue]{hyperref}

\def\wacvPaperID{1250} % *** Enter the WACV Paper ID here
\def\confName{WACV}
\def\confYear{2027}

\title{Soft Mixture-of-Recursions: Going Deeper with Recursive Vision Transformers}

\author{Sang In Lee \qquad Jihun Park\\
Chungnam National University, South Korea\\
{\tt\small sangin.lee.life@o.cnu.ac.kr}, \quad {\tt\small jihun.park@cnu.ac.kr}
}

\begin{document}
\maketitle
\input{sec/0_abstract}

\input{sec/1_intro}
\input{sec/2_relatedwork}
\input{sec/3_method}
\input{sec/4_experiments}
\input{sec/5_conclusion}

\section*{Acknowledgments}
This work was supported by the Institute of Information \& Communications Technology Planning \& Evaluation (IITP) grant funded by the Korea government (MSIT) (IITP-2026-RS-2026-25616664, AI Star Fellowship Support Program).

{
    \small
    \bibliographystyle{ieeenat_fullname}
    \bibliography{main}
}

\clearpage
\appendix

% \twocolumn[{
% \begin{center}
% {\Large \bf Soft Mixture-of-Recursions: Going Deeper with Recursive Vision Transformers\par}
% \vspace{0.5em}
% {\large Supplementary Material\par}
% \end{center}
% \vspace{1.0em}
% }]

\input{sec/supp}

\end{document}

%% file: preamble.tex
\newcommand{\red}[1]{{\color{red}#1}}

%% file: sec/0_abstract.tex
\begin{abstract}
Vision Transformers are typically scaled by increasing depth or width, at the cost of more unique parameters. We explore recursion as an alternative scaling axis that increases effective computational depth through parameter reuse. Yet simply increasing recursion depth does not reliably improve performance because representations from different recursion stages are not fully utilized. We propose Soft Mixture-of-Recursions (SoftMoR), which learns token-wise weights to combine representations across recursion steps, and instantiate it as the Soft Recursive Vision Transformer (SR-ViT). Experiments on ImageNet-1K, COCO, and ADE20K show that SR-ViT can benefit from deeper recursion while adding few parameters. On ImageNet-1K, increasing recursion depth from $T{=}1$ to $T{=}3$ improves top-1 accuracy from 79.83\% to 82.45\% with only 1.1M additional parameters. At matched effective depth, SR-ViT also outperforms independently parameterized counterparts with substantially fewer parameters. These results support recursive deepening as an effective, parameter-efficient scaling axis for Vision Transformers. The code is available at \small\url{https://github.com/sangin-lee/sr-vit}.
\end{abstract}

%% file: sec/1_intro.tex
\section{Introduction}
\label{sec:1}

\begin{figure}
    \centering
    \includegraphics[width=1.0\linewidth]{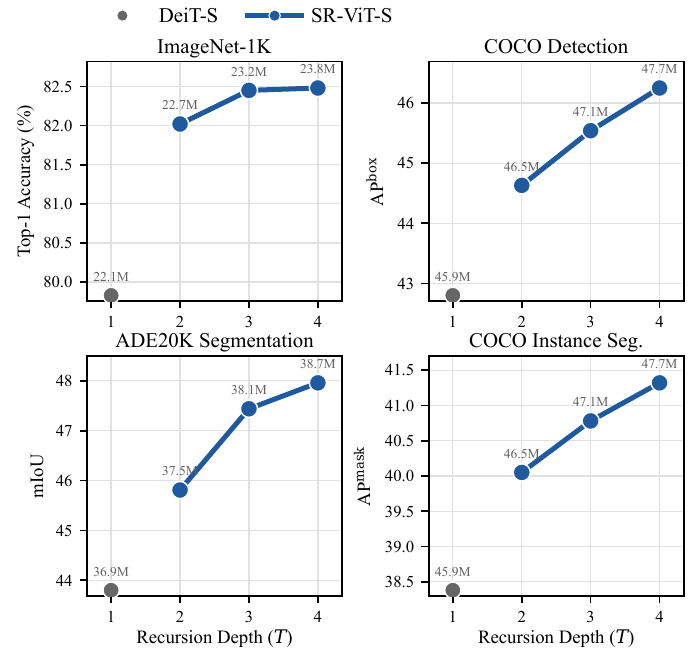}
    \vspace{-20pt}
    \caption{
    SoftMoR enables parameter-efficient recursive deepening of Vision Transformers. $T{=}1$ denotes the non-recursive DeiT-S baseline. Increasing $T$ expands the effective computational depth with only a small increase in unique parameters, and SR-ViT shows favorable scaling behavior across classification and dense prediction tasks.
    }
    \label{fig:1}
    \vspace{-15pt}
\end{figure}

Vision Transformers (ViTs)~\cite{dosovitskiy2020image} have demonstrated remarkable scalability across model size and training data, achieving strong performance on a wide range of computer vision tasks~\cite{he2022masked,li2022exploring,kirillov2023segment}. This scalability has been largely driven by increasingly larger architectures, ranging from compact models with only a few million parameters~\cite{mehta2022mobilevit,touvron2021training} to large-scale variants containing hundreds of millions or even billions of parameters~\cite{dosovitskiy2020image,zhai2022scaling,dehghani2023scaling}. While effective, such scaling entails substantial growth in parameter count, together with increased computational and memory requirements, leading to higher training and deployment costs. This motivates the exploration of alternative ways to increase the representational capacity of Vision Transformers without relying on increasingly larger parameterized architectures.

Among these directions, recent studies have increasingly focused on recursive Transformer architectures~\cite{shen2022sliced,bae2025relaxed,bae2025mor,he2026visionmor,li2026edgerecvit,yu2025mesh,xu2026looping}. Rather than using a separate set of parameters for each Transformer layer, these architectures reuse the same module across multiple computation steps, enabling deep computation with substantially fewer unique parameters. Existing studies have largely leveraged this property to construct compact, parameter-efficient models that remain competitive with non-recursive counterparts~\cite{bae2025relaxed,li2026edgerecvit}.

In particular, Mixture-of-Recursions (MoR)~\cite{bae2025mor} has recently drawn attention in autoregressive language modeling as a framework that combines recursive parameter sharing with token-wise adaptive computation. By assigning different recursion depths to different tokens, MoR allows the model to spend more computation on tokens that require deeper processing and fewer recursive steps on simpler ones. This formulation treats recursion depth primarily as a token-wise adaptive-computation decision. Since autoregressive inference proceeds token by token, such adaptive depth allocation is naturally aligned with language generation, where the computational demand may vary for each generated token.

For Vision Transformers, however, the Transformer is used as a backbone feature extractor. Tokens are processed jointly to form a spatially coherent representation, rather than being generated one at a time as prediction targets. Although token-wise adaptive computation can also be applied to ViTs, its benefit is less direct in this setting, since reducing computation for selected tokens may affect the quality of the overall visual representation, especially for dense prediction tasks that require spatially detailed features~\cite{liu2024revisiting}. We instead explore recursion from a different perspective: as a scaling axis for Vision Transformers. By reusing shared Transformer blocks across recursion steps, recursion can increase effective computational depth without a proportional increase in the number of unique parameters. This is because the output of each recursion step is passed through the same layers again at the next step, increasing the number of sequential layer applications along the computational path. The key question is therefore whether this additional recursive depth can be effectively translated into stronger visual representations.

However, a deeper recursive computation path does not automatically yield stronger representations. Our empirical study shows that simply applying more recursion can help at shallow recursion depths but does not reliably benefit from further increases in depth. We observe a similar limitation with MoR-style hard depth selection: although each token may choose a different recursion depth, only one representation is retained, while the others are discarded. This reduces the sequence of representations produced across multiple recursion steps to a single state. We hypothesize that this limitation makes it difficult to fully exploit the additional computational depth introduced by recursion, because useful information may emerge at different recursion stages along the computational trajectory, but such hard selection restricts access to these intermediate states.

To address this, we propose \emph{Soft Mixture-of-Recursions} (SoftMoR), which learns token-wise mixture weights and combines representations from all recursion steps. Rather than committing each token to a single point along the recursive trajectory, SoftMoR allows it to draw on multiple representations from earlier to later stages at different effective depths according to their learned relevance. In this way, information from different effective depths can directly contribute to the final representation, enabling the model to better utilize the computational depth provided by recursion.

We instantiate SoftMoR in a Vision Transformer as the \emph{Soft Recursive Vision Transformer} (SR-ViT). SR-ViT reuses standard ViT layers across recursion steps, so increasing recursion depth expands the effective computational depth with only a small increase in unique parameters. Together with SoftMoR, this provides a practical way to explore recursion as a scaling axis for Vision Transformers.

Extensive experiments across image classification, object detection, instance segmentation, and semantic segmentation demonstrate the effectiveness and generality of the proposed approach. \cref{fig:1} highlights the scaling behavior of SR-ViT-S. Here, $T$ denotes the recursion depth. Starting from $T{=}1$, where no recursive reuse is performed, increasing the recursion depth to $T{=}2$, $3$, and $4$ progressively expands the effective computational depth. 
SR-ViT shows favorable scaling behavior with increasing $T$ across the evaluated tasks, with only minimal parameter overhead.
On ImageNet-1K, increasing $T$ from 1 to 2 and 3 raises top-1 accuracy from 79.83\% to 82.02\% and 82.45\%, while adding only 1.1M parameters. At this nearly fixed parameter budget, allocating more computation through recursion outperforms substantially larger standard ViTs, supporting recursive deepening as a parameter-efficient alternative to conventional parameter scaling.

Our contributions are summarized as follows:
\begin{itemize}
    \item We propose SoftMoR and SR-ViT, a Vision Transformer architecture that recursively reuses Transformer blocks and softly combines representations from multiple recursion stages to better exploit increased computational depth.

    \item We demonstrate that SR-ViT exhibits favorable scaling behavior with recursion depth across diverse vision tasks, including image classification, object detection, instance segmentation, and semantic segmentation.

    \item Our results show that recursion can expand effective computational depth while keeping the unique parameter count nearly unchanged, supporting recursive deepening as a useful, parameter-efficient scaling axis for Vision Transformers.
\end{itemize}

%% file: sec/2_relatedwork.tex
\section{Related Work}
\label{sec:2}

\subsection{Deeper and Recursive Transformers}

Increasing Transformer depth has been studied as a way to improve representation capacity, but directly stacking more Vision Transformer blocks does not always yield consistent gains. DeepViT observed that very deep ViTs suffer from attention collapse and proposed re-attention to increase attention diversity across layers~\cite{zhou2021deepvit}. CaiT improved the optimization of deeper image Transformers through LayerScale and class-attention, enabling deeper models to benefit more effectively from added layers~\cite{touvron2021going}. These methods increase depth by adding independently parameterized Transformer layers, while our work scales network depth by recursively reusing shared blocks.

Recursive and looped Transformers provide another path to deeper computation by reusing shared modules across multiple computation steps. The Universal Transformer introduced depth-wise recurrence with adaptive per-position halting~\cite{dehghani2018universal}. Related parameter-sharing perspectives have also been explored in language models, such as cross-layer sharing in ALBERT~\cite{lan2019albert} and equilibrium-style weight-tied computation in Deep Equilibrium Models~\cite{bai2019deep}. In vision, SReT applied recursive weight sharing to ViTs and introduced sliced self-attention to reduce the computation overhead of repeated block application~\cite{shen2022sliced}. More recent language models have explored relaxed recursive sharing with layer-wise LoRA~\cite{bae2025relaxed}, persistent memory across recursive steps~\cite{yu2025mesh}, and elastic-depth looped computation for latent reasoning~\cite{jeddi2026loopformer}. These studies demonstrate the potential of shared recursive computation, often with an emphasis on compact model design, memory efficiency, or variable-depth reasoning. In contrast, our work studies recursion as a scaling axis for increasing effective computational depth, rather than primarily as a mechanism for model compression or variable-depth reasoning.

\subsection{Adaptive ViTs and Mixture-of-Recursions}

Adaptive computation methods aim to allocate different amounts of computation to different inputs or tokens. In Vision Transformers, several works reduce computation by dynamically selecting or reorganizing image tokens. DynamicViT prunes redundant tokens through learned token sparsification~\cite{rao2021dynamicvit}, A-ViT applies adaptive halting to spatial tokens~\cite{yin2022avit}, EViT preserves attentive tokens while fusing inattentive ones~\cite{liang2022evit}, and ATS adaptively samples informative tokens using attention-based scores~\cite{fayyaz2022ats}. These methods mainly improve ViT efficiency by reducing token-level computation.

Mixture-of-Recursions (MoR) combines recursive parameter sharing with token-wise adaptive computation in autoregressive language models~\cite{bae2025mor}. A learnable router assigns each token a recursion depth according to its required processing, described as an adaptive depth of token-level ``thinking.'' By processing only the tokens that remain active at each recursion step and selectively caching their key-value states, MoR jointly targets parameter, computation, and memory efficiency. Vision-MoR applies this formulation to image patches using a shifted-window-based spatial router and patch-wise early exiting~\cite{he2026visionmor}. Closely related, Edge-RecViT employs an MLP-based ranker to determine token-specific recursion depths~\cite{li2026edgerecvit}. Both vision approaches are primarily framed around adaptive computation and efficient model design, selectively processing informative tokens more deeply while allowing other tokens to exit earlier. In contrast, our work does not halt tokens at different depths for efficiency. We apply a given recursion depth to the full representation and softly combine all recursive-step outputs, so that increasing recursion depth expands the effective computational depth while allowing representations from multiple recursion steps to contribute to the output.

%% file: sec/3_method.tex
\begin{figure*}[t]
    \centering
    \includegraphics[width=1\linewidth]{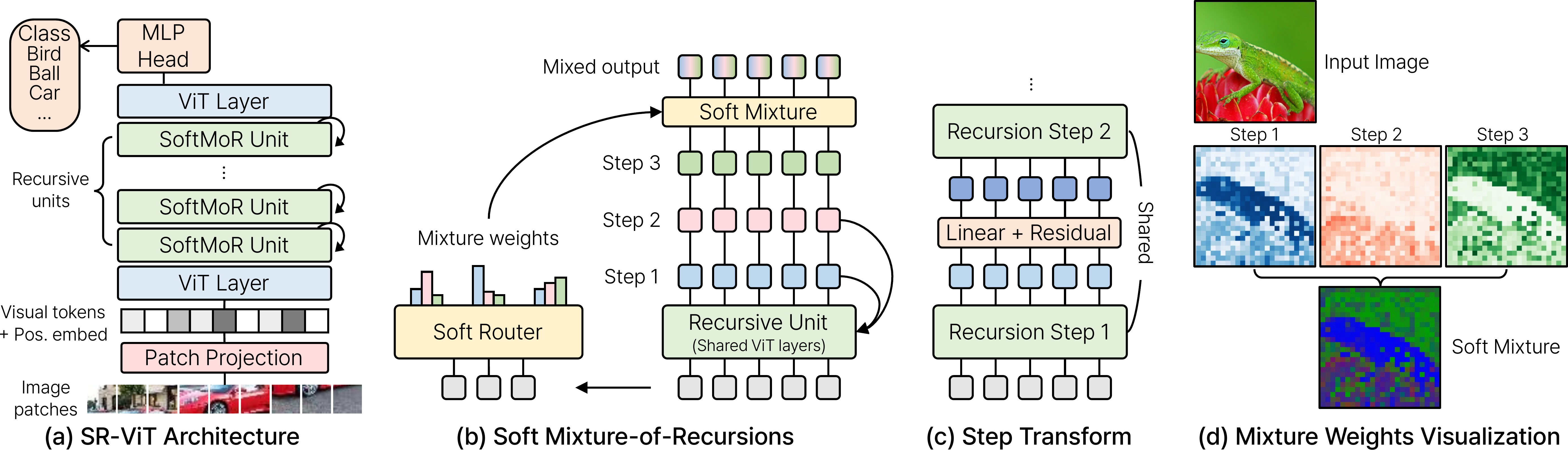}
    \caption{\textbf{Overview of Soft Recursive Vision Transformer (SR-ViT) with Soft Mixture-of-Recursions (SoftMoR).} (a) SR-ViT organizes intermediate ViT layers into SoftMoR units, where shared Transformer blocks are recursively reused across computation steps. (b) SoftMoR learns token-wise mixture weights over recursion steps using the \emph{Soft Router} and aggregates outputs from all recursion steps through a \emph{Soft Mixture}. (c) A \emph{Step Transform} is inserted between consecutive recursion steps to encourage step-specific variation across recursive computation. (d) Learned mixture weights show that different spatial locations utilize recursive representations from different recursion steps. The blended map illustrates the resulting Soft Mixture by combining the step-wise weight maps.}
    \label{fig:3}
\end{figure*}

\section{Method}
\label{sec:3}

We first formulate recursive computation in Vision Transformers and describe how parameter reuse increases computational depth. We then introduce Soft Mixture-of-Recursions (SoftMoR), which learns token-wise soft mixtures over the representations produced at all recursion steps. A lightweight step transform further encourages variation among recursive representations generated with shared Transformer parameters. Finally, we instantiate SoftMoR in a standard Vision Transformer architecture, yielding the Soft Recursive Vision Transformer (SR-ViT).

\subsection{Recursive Vision Transformer}
\label{sec:3.1}

\paragraph{Standard Vision Transformer.}
Given an input token sequence $\mathbf{X}_0 \in \mathbb{R}^{N \times D}$ with $N$ tokens and embedding dimension $D$, a standard Vision Transformer sequentially applies independent Transformer blocks:
\begin{equation}
    \mathbf{X}_{\ell+1} = F_{\ell}(\mathbf{X}_{\ell}),
\end{equation}
where $F_{\ell}(\cdot)$ denotes the $\ell$-th Transformer block composed of multi-head self-attention and feed-forward layers with residual connections.

A Vision Transformer with $L$ Transformer blocks can therefore be written as
\begin{equation}
    \mathbf{X}_{L}
    =
    F_{L-1}
    \circ
    F_{L-2}
    \circ
    \cdots
    \circ
    F_{0}
    (\mathbf{X}_{0}),
\end{equation}
where each block is independently parameterized.

\paragraph{Recursive Transformer.}
Unlike standard Vision Transformers, recursive Transformers repeatedly apply a shared recursive unit across multiple recursion steps. The recursive unit may consist of one or more Transformer blocks and is denoted by $\mathcal{R}_{\Theta}(\cdot)$, where $\Theta$ denotes the parameters shared across recursion steps. Given an initial representation $\mathbf{X}^{(0)}=\mathbf{X}$, the recursive unit is applied repeatedly as
\begin{equation}
    \mathbf{X}^{(t+1)}
    =
    \mathcal{R}_{\Theta}(\mathbf{X}^{(t)}),
    \quad t = 0, \dots, T-1,
\end{equation}
where $T$ denotes the recursion depth, i.e., the number of times the recursive unit is applied.

After $T$ recursive applications, the resulting representation can be written as:
\begin{equation}
    \mathbf{X}^{(T)}
    =
    \underbrace{
    \mathcal{R}_{\Theta}
    \circ
    \mathcal{R}_{\Theta}
    \circ
    \cdots
    \circ
    \mathcal{R}_{\Theta}
    }_{T \text{ recursion steps}}
    (\mathbf{X}^{(0)}),
\end{equation}
or equivalently,
\begin{equation}
    \mathbf{X}^{(T)}
    =
    \mathcal{R}_{\Theta}^{T}
    (\mathbf{X}^{(0)}).
\end{equation}

A recursive unit containing $K$ sequential Transformer blocks is defined as
\begin{equation}
\mathcal{R}_{\Theta}
=
F_{\theta_K}
\circ
F_{\theta_{K-1}}
\circ
\cdots
\circ
F_{\theta_1},
\end{equation}
where $\theta_k$ denotes the parameters of the $k$-th Transformer block and
\(
\Theta = \{\theta_1, \theta_2, \dots, \theta_K\}
\)
denotes the parameter set of the recursive unit. The output after $T$ recursion steps can be expressed as
\begin{equation}
    \mathbf{X}^{(T)}
    =
    \left(
    F_{\theta_K}
    \circ
    F_{\theta_{K-1}}
    \circ
    \cdots
    \circ
    F_{\theta_1}
    \right)^T
    (\mathbf{X}^{(0)}).
\end{equation}

We use \emph{effective computational depth} to denote the number of Transformer block applications along the unrolled computation path, rather than the number of independently parameterized layers. For a single recursive unit containing $K$ Transformer blocks, applying the unit for $T$ recursion steps gives
\begin{equation}
D_{\mathrm{eff}} = K \times T.
\end{equation}
Increasing $T$ therefore increases computational depth while keeping the number of unique Transformer parameters fixed. For architectures containing multiple recursive units and standard non-recursive blocks, the total effective depth is obtained by counting all Transformer block applications along the unrolled computation path.

\subsection{Soft Mixture-of-Recursions}
\label{sec:3.2}

We propose \emph{Soft Mixture-of-Recursions} (SoftMoR), which learns token-wise mixture weights to aggregate representations from all recursion steps. Unlike existing Mixture-of-Recursions formulations that assign each token to a single recursion depth through hard routing, SoftMoR allows each token to combine representations from multiple effective depths.

\paragraph{Soft Mixture.}
Applying a recursive unit for $T$ recursion steps produces a sequence of recursive representations
\(
{\mathbf{X}^{(1)}, \mathbf{X}^{(2)}, \dots, \mathbf{X}^{(T)}},
\)
where $\mathbf{X}^{(t)}$ denotes the output after the $t$-th recursion step. As illustrated in \cref{fig:3}(b), SoftMoR employs a lightweight linear router that maps each input token representation to $T$ mixture logits, one for each recursion step:
\begin{equation}
\mathbf{Z}
=
G(\mathbf{X}^{(0)})
\in
\mathbb{R}^{N \times T},
\end{equation}
where $G:\mathbb{R}^{D}\rightarrow\mathbb{R}^{T}$ is applied independently to each token. The mixture weights are obtained by applying a softmax over the recursion-step dimension:
\begin{equation}
\alpha_n^{(t)}
=
\frac{
\exp\left(Z_n^{(t)}\right)
}{
\sum_{k=1}^{T}
\exp\left(Z_n^{(k)}\right)
},
\end{equation}
where $\alpha_n^{(t)}$ denotes the mixture weight assigned to recursion step $t$ for the $n$-th token.

The final token representation is obtained by softly aggregating the representations from all recursion steps:
\begin{equation}
\mathbf{X}^{\mathrm{out}}_n
=
\sum_{t=1}^{T}
\alpha_n^{(t)}
\mathbf{X}^{(t)}_n.
\end{equation}
Thus, each token forms its final representation by combining outputs from all recursion steps using mixture weights predicted from its input representation. As visualized in \cref{fig:3}(d), the learned mixture weights form spatially coherent patterns, and the blended map illustrates how Soft Mixture combines step-wise contributions.

\begin{figure*}[t]
    \centering

    \begin{minipage}[t]{0.64\linewidth}
        \vspace{0pt}
        \centering
        \includegraphics[width=\linewidth]{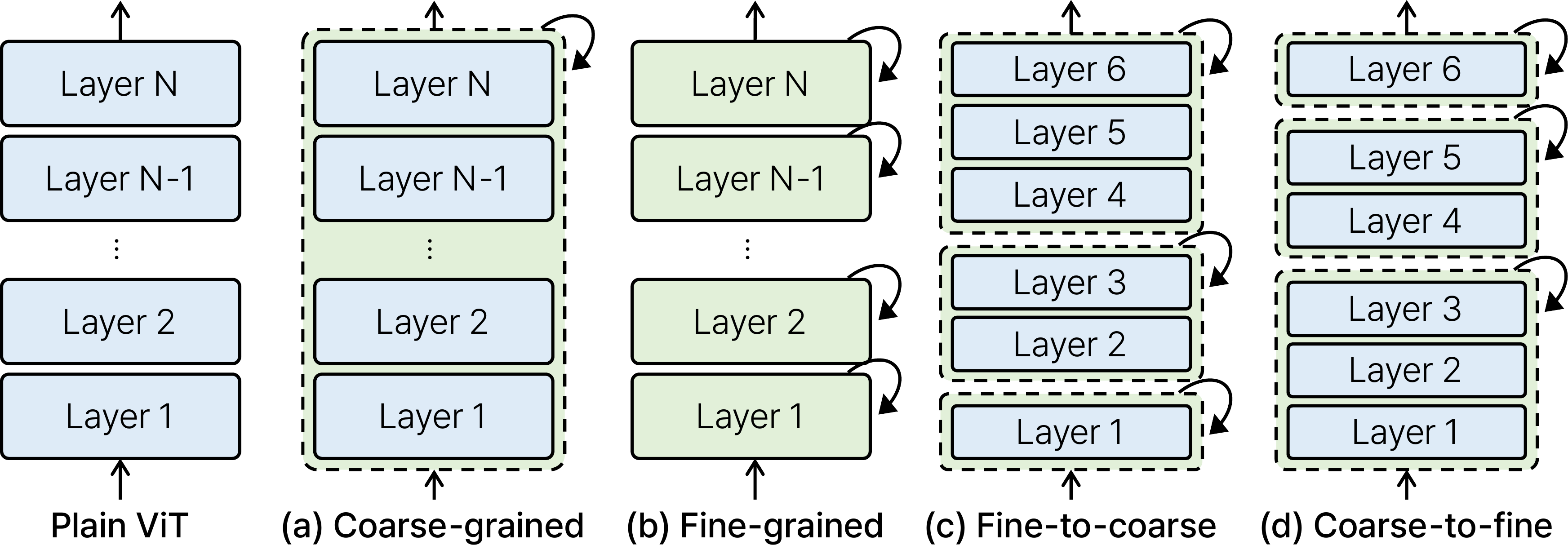}
        \vspace{-18pt}
        \caption{Conceptual illustrations of different recursive unit organizations. Blue boxes denote standard ViT layers, while green regions denote recursive units.}
        \label{fig:recursion_unit}
    \end{minipage}
    \hfill
    \begin{minipage}[t]{0.33\linewidth}
        % \vspace{20pt}
        \centering
        \captionof{table}{ImageNet-1K results for different recursive unit organizations at $T{=}2$.}
        \begin{tabular}{lcc}
        \toprule
        Model & Top-1 (\%) \\
        \midrule
        DeiT-S (non-recursive) & 79.83 \\
        (a) Coarse-grained & 80.48 \\
        (b) Fine-grained & 80.33 \\
        (c) Fine-to-coarse & 78.89 \\
        (d) Coarse-to-fine & \textbf{81.12} \\
        \bottomrule
        \end{tabular}

        \vspace{0.5em}
        \label{tab:recursion_unit_acc}
    \end{minipage}
    \vspace{-10pt}
\end{figure*}

\paragraph{Step Transform.}
The Transformer parameters within each recursive unit are shared across recursion steps, which may limit the flexibility of representations produced at different depths. To encourage step-specific variation, we insert a lightweight \emph{Step Transform} between consecutive recursion steps, as illustrated in \cref{fig:3}(c). Each transition applies an independently parameterized token-wise linear projection with a residual connection, and the transformed representation is passed to the next recursion step.

\paragraph{Soft Recursive Vision Transformer.}
As illustrated in \cref{fig:3}(a), we instantiate SoftMoR in a Vision Transformer to construct the \emph{Soft Recursive Vision Transformer} (SR-ViT). SR-ViT preserves the standard ViT components and Transformer block design, while organizing intermediate Transformer blocks into recursive units whose parameters are reused across recursion steps. Within each recursive unit, SoftMoR aggregates the representations produced across recursion steps, with step transforms applied between consecutive steps. We empirically determine the organization of these recursive units in \cref{sec:4.1}.

%% file: sec/4_experiments.tex
\begin{table}[h]
\centering
\caption{SR-ViT specifications. Blocks denotes unique Transformer blocks; $D_{\mathrm{eff}}$ is the resulting sequential depth.}
\label{tab:model-specs}
\resizebox{\linewidth}{!}{
\begin{tabular}{lccccc}
\toprule
Model & Blocks & $D_{\mathrm{eff}}$ & Dim & Params & FLOPs \\
\midrule
SR-ViT-T ($T{=}2$) & 8 & 14 & 384 & 15.4M & 5.4G \\
SR-ViT-T ($T{=}3$) & 8 & 20 & 384 & 15.8M & 7.8G \\
\midrule
SR-ViT-S ($T{=}2$) & 12 & 22 & 384 & 22.7M & 8.5G \\
SR-ViT-S ($T{=}3$) & 12 & 32 & 384 & 23.2M & 12.4G \\
SR-ViT-S ($T{=}4$) & 12 & 42 & 384 & 23.8M & 16.3G \\
\midrule
SR-ViT-B ($T{=}2$) & 12 & 22 & 512 & 39.9M & 14.8G \\
SR-ViT-B ($T{=}3$) & 12 & 32 & 512 & 41.0M & 21.6G \\
\bottomrule
\end{tabular}
}
\end{table}

\section{Experiments}

\begin{table*}[!t]
\centering
\caption{
ImageNet-1K classification results under standard DeiT training and DeiT-style distillation. Top-1 values marked with \textdagger~use DeiT-style hard distillation with a distillation token and teacher supervision.
}
\label{tab:imagenet-results}
\resizebox{\textwidth}{!}{
\begin{tabular}{lcccccccccc}
\toprule

&
&
&
\multicolumn{4}{c}{\textbf{Resolution: 224 $\times$ 224}} &
\multicolumn{4}{c}{\textbf{Resolution: 384 $\times$ 384}} \\

\cmidrule(r){4-7}
\cmidrule(l){8-11}

Model &
Params &
$D_{\mathrm{eff}}$ &
FLOPs &
img/s &
Top-1 (\%) &
Top-1\textsuperscript{\textdagger} (\%) &
FLOPs &
img/s &
Top-1 (\%) &
Top-1\textsuperscript{\textdagger} (\%) \\

\midrule

DeiT-S~\cite{touvron2021training} & 22.1M & 12 & 4.6G & 2614 & 79.83 & 81.18 & 15.5G & 750 & -- & -- \\
DeiT-B~\cite{touvron2021training} & 86.6M & 12 & 17.5G & 881 & 81.80 & 83.33 & 55.5G & 264 & 83.10 & 84.50 \\

\midrule \addlinespace[3pt]

SR-ViT-T ($T{=}2$) & 15.4M & 14 & 5.4G & 2121 & 79.89 & 81.03 & 18.3G & 617 & 82.21 & 83.12 \\
SR-ViT-T ($T{=}3$) & 15.8M & 20 & 7.8G & 1477 & 80.93 & 81.89 & 26.2G & 432 & 83.22 & 83.73 \\

\midrule \addlinespace[3pt]

SR-ViT-S ($T{=}2$) & 22.7M & 22 & 8.5G & 1370 & 82.02 & 82.67 & 28.6G & 396 & 84.07 & 84.27 \\
SR-ViT-S ($T{=}3$) & 23.2M & 32 & 12.4G & 943 & 82.45 & 83.36 & 41.7G & 273 & 84.19 & 84.87 \\
SR-ViT-S ($T{=}4$) & 23.8M & 42 & 16.3G & 718 & 82.48 & 83.57 & 54.8G & 210 & 84.24 & 85.08 \\

\midrule \addlinespace[3pt]

SR-ViT-B ($T{=}2$) & 39.9M & 22 & 14.8G & 898 & \textbf{82.76} & 83.75 & 48.3G & 260 & \textbf{84.39} & 85.07 \\
SR-ViT-B ($T{=}3$) & 41.0M & 32 & 21.6G & 620 & 82.19 & \textbf{84.02} & 70.4G & 187 & 83.53 & \textbf{85.25} \\

\bottomrule
\end{tabular}
}
\vspace{-7pt}
\end{table*}

\subsection{Does Recursion Improve Vision Transformers?}
\label{sec:4.1}
Before evaluating SoftMoR, we first ask a more fundamental question: can recursion itself improve Vision Transformer representations? Although recursion increases effective computational depth, it is not guaranteed that repeatedly applying the same shared blocks will produce stronger visual representations. We therefore study naive recursive ViTs, without soft mixture or step transforms, to isolate the effect of recursion and determine the recursive unit organization used in SR-ViT.

Following prior recursive Transformer designs~\cite{bae2025mor,yu2025mesh,li2026edgerecvit}, we keep the initial and final Transformer blocks non-recursive and apply recursion to the intermediate blocks. However, the same intermediate blocks can be grouped into recursive units in different ways~\cite{bae2025mor}. The entire stack can be grouped into a single coarse-grained recursive unit~\cite{bae2025relaxed}, or each Transformer block can form a separate fine-grained recursive unit~\cite{shen2022sliced}. We also consider depth-dependent organizations in which the unit granularity changes across network depth, resulting in fine-to-coarse and coarse-to-fine variants. To study the effect of this design choice, we compare the recursive unit organizations illustrated in \cref{fig:recursion_unit}. All configurations are implemented on DeiT-S with recursion depth $T{=}2$, and differ only in how the intermediate Transformer blocks are grouped into recursive units.

\cref{tab:recursion_unit_acc} summarizes the results. Recursion can improve Vision Transformer performance, but its effectiveness depends strongly on the recursive unit organization. Coarse-grained and fine-grained organizations provide only modest gains over the DeiT-S baseline, while the fine-to-coarse strategy fails to improve performance. In contrast, the coarse-to-fine organization achieves the strongest result, improving ImageNet-1K top-1 accuracy from 79.83\% to 81.12\%. We therefore adopt the coarse-to-fine organization for all SR-ViT variants. Further analysis of recursive-unit organizations from a representation perspective is provided in Appendix \cref{sec:supp_org}.

Nevertheless, the improvement from naive recursion remains limited relative to the additional recursive computation. Although recursion improves performance at $T{=}2$, deeper naive recursion does not reliably translate into additional gains, as further analyzed in \cref{sec:4.6}. This observation suggests that naive recursive scaling alone is insufficient for effectively exploiting deeper recursive computation.

\subsection{Model Variants}
\label{sec:4.2}

Following \cref{sec:4.1}, the first and last Transformer blocks remain non-recursive, and the remaining blocks are organized in a coarse-to-fine layout and applied for $T$ recursion steps. As shown in \cref{tab:model-specs}, increasing $T$ therefore raises the effective computational depth $D_{\mathrm{eff}}$ and the corresponding FLOPs, while the number of unique Transformer parameters stays fixed. Detailed architectural specifications, including recursive-unit layouts, are provided in Appendix \cref{sec:supp_specs}.

\subsection{ImageNet-1K Classification}
\label{sec:4.3}

We evaluate SR-ViT on ImageNet-1K~\cite{deng2009imagenet} using the standard DeiT training protocol~\cite{touvron2021training}. All models are trained from scratch for 300 epochs with the same optimization, augmentation, and regularization settings as DeiT, without architecture-specific training tricks or additional hyperparameter tuning. For DeiT-style distillation, we follow the original DeiT setup using a RegNetY-16GF~\cite{radosavovic2020designing} teacher with hard distillation. Results are reported in \cref{tab:imagenet-results}.

% As shown in \cref{tab:imagenet-results}, \rev{SR-ViT exhibits overall favorable scaling behavior with increasing recursion depth. For example, at $224\times224$, SR-ViT-S uses a parameter budget close to DeiT-S (22.1M). Applying more computation through recursion raises accuracy from 79.83\% to 82.02\% at $T{=}2$ and 82.45\% at $T{=}3$, with only a marginal further gain at $T{=}4$ (82.48\%). \rev{This depth-compute scaling compares favorably with conventional width scaling in this setting}: SR-ViT-S surpasses DeiT-B (81.80\%, 86.8M, 17.5G) at $T{=}2$ with less than half the FLOPs, and remains more accurate at $T{=}3$ while still using fewer FLOPs.} SR-ViT also remains strong at higher resolution, where SR-ViT-T ($T{=}3$, 15.8M) surpasses DeiT-B.

As shown in \cref{tab:imagenet-results}, SR-ViT exhibits favorable scaling behavior with increasing recursion depth. At $224\times224$, SR-ViT-S uses a parameter budget near DeiT-S (22.1M). Applying more computation through recursion raises accuracy from 79.83\% to 82.02\% at $T{=}2$ and 82.45\% at $T{=}3$, with little gain at $T{=}4$ (82.48\%). This depth-compute scaling compares favorably with conventional width scaling here: SR-ViT-S surpasses DeiT-B (81.80\%, 86.6M, 17.5G) at $T{=}2$ with less than half the FLOPs, and remains more accurate at $T{=}3$ while using fewer FLOPs. At higher resolution, SR-ViT-T ($T{=}3$, 15.8M) also surpasses DeiT-B.

As reported in \cref{tab:imagenet-results}, increasing $T$ raises $D_{\mathrm{eff}}$, and the associated FLOPs increase and throughput decrease match what is expected from a sequentially deeper ViT with the increased depth. Further analysis on efficiency, including training and inference cost, is provided in Appendix \cref{sec:supp_efficiency}.

DeiT-style distillation further improves SR-ViT across model scales and recursion depths, showing that the proposed architecture remains fully compatible with standard ViT training paradigms. With distillation at 384 resolution, the strongest SR-ViT models exceed 85\% top-1 accuracy, showing that scaling depth through recursion can reach highly competitive classification performance.

\begin{table*}[t]
\centering
\caption{Downstream transfer results on COCO 2017 object detection/instance segmentation and ADE20K semantic segmentation. COCO uses Mask R-CNN and ADE20K uses UPerNet. Params and FLOPs denote the full downstream model, including task-specific heads.}
\label{tab:coco-ade20k-results}
\resizebox{\textwidth}{!}{
\begin{tabular}{lcccccccc|cccc}
\toprule
\multirow{3}{*}{Backbone}
& \multicolumn{8}{c}{\textbf{COCO val}} \vline
& \multicolumn{4}{c}{\textbf{ADE20K val}} \\
\cmidrule(lr){2-9}
\cmidrule(lr){10-13}
&
\multirow{2}{*}{Params}
& \multirow{2}{*}{FLOPs}
& \multicolumn{3}{c}{Box AP}
& \multicolumn{3}{c}{Mask AP} \vline
& \multirow{2}{*}{Params}
& \multirow{2}{*}{FLOPs}
& \multicolumn{2}{c}{mIoU} \\
\cmidrule(lr){4-6}
\cmidrule(lr){7-9}
\cmidrule(lr){12-13}
&
&
&
AP$^{b}$
& AP$_{50}^{b}$
& AP$_{75}^{b}$
& AP$^{m}$
& AP$_{50}^{m}$
& AP$_{75}^{m}$
&
&
&
s.s.
& m.s. \\
\midrule
DeiT-S~\cite{touvron2021training}
& 45.9M
& 378.5G
& 42.80
& 64.94
& 46.05
& 38.38
& 61.58
& 40.57
& 36.9M
& 232.2G
& 43.80
& 45.07
\\
DeiT-B~\cite{touvron2021training}
& 113.8M
& 654.6G
& 47.07
& 69.09
& 51.37
& 42.12
& 65.83
& 45.17
& 121.5M
& 564.4G
& 45.87
& 46.25
\\
\midrule
SR-ViT-S ($T{=}2$)
& 46.5M
& 453.6G
& 44.63
& 66.75
& 48.04
& 40.05
& 63.21
& 42.63
& 37.5M
& 285.9G
& 45.81
& 46.95
\\
SR-ViT-S ($T{=}3$)
& 47.1M
& 528.8G
& 45.54
& 67.68
& 49.45
& 40.78
& 64.29
& 43.72
& 38.1M
& 339.6G
& 47.44
& 48.79
\\
SR-ViT-S ($T{=}4$)
& 47.7M
& 603.9G
& 46.25
& 68.29
& 50.68
& 41.32
& 64.94
& 44.30
& 38.7M
& 393.2G
& \textbf{47.96}
& \textbf{49.05}
\\
\midrule
SR-ViT-B ($T{=}2$)
& 64.7M
& 583.9G
& 45.87
& 67.94
& 49.77
& 41.12
& 64.55
& 43.92
& 60.2M
& 408.4G
& 47.31
& 48.80
\\
SR-ViT-B ($T{=}3$)
& 65.8M
& 717.4G
& \textbf{47.15}
& \textbf{69.38}
& \textbf{51.65}
& \textbf{42.17}
& \textbf{65.98}
& \textbf{45.36}
& 61.2M
& 496.6G
& 47.16
& 48.35
\\
\bottomrule
\end{tabular}
}
\end{table*}

\subsection{COCO Detection and Instance Segmentation}
\label{sec:4.4}

We evaluate the transferability of SR-ViT representations on COCO 2017~\cite{lin2014microsoft} for object detection and instance segmentation. Following ViTDet~\cite{li2022exploring}, ImageNet-pretrained backbones are integrated into Mask R-CNN~\cite{he2017mask} using a simple feature pyramid constructed from the final backbone feature map. All models are fine-tuned under the same training protocol using Detectron2~\cite{wu2019detectron2}, including Large Scale Jittering (LSJ)~\cite{ghiasi2021simple} with $1024\times1024$ training crops and the $3\times$ schedule, with the backbone architecture being the only variable. The COCO results are reported in the left part of \cref{tab:coco-ade20k-results}.

The COCO results show that scaling through recursion also works for object detection and instance segmentation. For SR-ViT-S, increasing the recursion depth consistently improves both box AP and mask AP, indicating that deeper recursive computation strengthens representations useful for spatial recognition tasks. This is important because COCO requires localized and spatially detailed features, rather than only image-level classification.

The larger SR-ViT-B variants show a similar trend. In particular, SR-ViT-B with $T{=}3$ slightly outperforms the DeiT-B counterpart in both box AP and mask AP, while using substantially fewer parameters in the full detection model. These results indicate that scaling the backbone through deeper recursion produces stronger visual representations, which transfer effectively to both object detection and instance segmentation.

% \begin{table}[t]
% \centering
% \label{tab:seg_results}
% \resizebox{\linewidth}{!}{
% \begin{tabular}{lccccc}
% \toprule
% \multirow{2}{*}{Backbone} &
% \multirow{2}{*}{Params} &
% \multirow{2}{*}{GFLOPs} &
% \multicolumn{2}{c}{mIoU} \\
% \cmidrule(lr){4-5}
% &
% &
% &
% s.s. &
% m.s. \\
% \midrule

% DeiT-S
% & 36.9M
% & 232.2
% & 43.80
% & 45.07
% \\

% DeiT-B
% & 121.5M
% & 564.4
% & 45.87
% & 46.25
% \\

% \midrule

% SR-ViT-S ($T{=}2$)
% & 37.5M
% & 285.9
% & 45.81
% & 46.95
% \\

% SR-ViT-S ($T{=}3$)
% & 38.1M
% & 339.6
% & 47.44
% & 48.79
% \\

% SR-ViT-S ($T{=}4$)
% & 38.7M
% & 393.2
% & \textbf{47.96}
% & \textbf{49.05}
% \\

% \midrule

% SR-ViT-B ($T{=}2$)
% & 60.2M
% & 408.4
% & 47.31
% & 48.80
% \\

% SR-ViT-B ($T{=}3$)
% & 61.2M
% & 496.6
% & 47.16
% & 48.35
% \\

% \bottomrule
% \end{tabular}
% }
% \caption{Semantic segmentation performance comparison on different backbones.}
% \end{table}

\subsection{ADE20K Semantic Segmentation}
\label{sec:4.5}

We further evaluate SR-ViT on ADE20K~\cite{zhou2017scene} semantic segmentation to assess its transferability to pixel-level dense prediction. All ImageNet-pretrained backbones are integrated into UPerNet~\cite{xiao2018unified} using MMSegmentation~\cite{mmseg2020} with a Multi-Level Neck (MLN), and trained under the same 160K iteration schedule with standard data augmentation and a $512\times512$ random crop. The ADE20K results are reported in the right part of \cref{tab:coco-ade20k-results}.

ADE20K provides a strong test of whether the recursive backbone produces spatial representations that remain useful for dense prediction. As shown in \cref{tab:coco-ade20k-results}, SR-ViT-S shows a clear and consistent improvement as the recursion depth increases. Both single-scale (s.s.) and multi-scale (m.s.) mIoU improve from $T{=}2$ to $T{=}3$, and further improve at $T{=}4$. This trend suggests that semantic segmentation can effectively benefit from deeper recursive computation, where pixel-level prediction requires rich and spatially detailed visual representations.

The comparison with DeiT-B further highlights this transfer behavior. Despite using a substantially smaller full segmentation model, SR-ViT-S with $T{=}3$ already outperforms DeiT-B under both single-scale and multi-scale evaluation. Further increasing the recursion depth to $T{=}4$ gives the best ADE20K performance among all evaluated models. These findings show that deeper recursive representations remain effective beyond classification, extending to pixel-level dense prediction.

\subsection{Analysis}
\label{sec:4.6}

\paragraph{Effect of Soft Mixture and Step Transform.}
We ablate the main components of SR-ViT in \cref{tab:ablation-depth}. Starting from the DeiT-S baseline, naive recursion improves performance at a shallow recursion depth, reaching 81.12\% top-1 accuracy at $T{=}2$. However, simply increasing the recursion depth does not guarantee further improvement. When $T$ is increased to 3, naive recursion drops to 79.17\%, falling below the non-recursive baseline. This shows that recursive reuse alone is not sufficient to reliably benefit from deeper computation in this setting. Using Step Transform without Soft Mixture is also insufficient: at $T{=}2$ it reaches 78.82\%, below the DeiT-S baseline.

Soft Mixture directly addresses this limitation by aggregating representations from all recursion steps. With Soft Mixture, performance improves over naive recursion at both recursion depths. Importantly, the model continues to improve when $T$ is increased from 2 to 3, suggesting that softly aggregating recursive outputs helps translate additional recursive computation into stronger representations. This shows that intermediate recursive representations contain useful information and should be explicitly utilized rather than discarded. Adding Step Transform further improves performance, reaching 82.02\% at $T{=}2$ and 82.45\% at $T{=}3$, with only a small increase in parameter count and FLOPs.
Simply adding layers does not improve over DeiT-S (see Sec.~E of the supp.). Applying Soft Mixture to those stacks yields only 80.53\% and 80.51\%, despite 39.8M / 57.6M parameters, and remains below recursive Soft Mixture (81.59\% / 81.74\%) at 22.1M parameters.
%These results indicate that recursive deepening is effective together with Soft Mixture and Step Transform: Soft Mixture makes additional recursive steps useful, while Step Transform provides additional flexibility across recursion steps.
These results indicate that increasing recursion depth alone is insufficient, while Soft Mixture helps convert additional recursive computation into useful representations and Step Transform provides flexibility across recursion steps.

\begin{table}[t]
\centering
\caption{
% Effect of the proposed components on SR-ViT-S under increasing recursion depth. Top-1 improvements are relative to DeiT-S.
% The bottom block applies Soft Mixture to independent layers at the same $D_{\mathrm{eff}}$.
Effect of the proposed components on SR-ViT-S.
Top-1 gains are relative to DeiT-S.
Non-recursive + Soft Mixture replaces recursively shared blocks with independent blocks at matched $D_{\mathrm{eff}}$/FLOPs, mixing features at corresponding depths.
}
\label{tab:ablation-depth}
\resizebox{\linewidth}{!}{
\begin{tabular}{lccc}
\toprule
Method & Params & FLOPs & Top-1 (\%) \\
\midrule
DeiT-S (Baseline)
& 22.1M
& 4.6G
& 79.83
\\
\midrule
Naive Recursion ($T{=}2$, $D_{\mathrm{eff}}{=}22$)
& 22.1M
& 8.4G
& 81.12 \footnotesize {($+$1.29)}
\\
\hspace{0.5em}w/ Step Transform only
& 22.7M
& 8.5G
& 78.82 \footnotesize \red{($-$1.01)}
\\
{\hspace{0.5em}w/ Soft Mixture only}
& 22.1M
& 8.4G
& 81.59 \footnotesize {($+$1.76)}
\\
{\hspace{0.5em}Full model}
& 22.7M
& 8.5G
& \textbf{82.02 \footnotesize {($+$2.19)}}
\\
\midrule
Naive Recursion ($T{=}3$, $D_{\mathrm{eff}}{=}32$)
& 22.1M
& 12.2G
& 79.17 \footnotesize \red{($-$0.66)}
\\
{\hspace{0.5em}w/ Soft Mixture only}
& 22.1M
& 12.2G
& 81.74 \footnotesize {($+$1.91)}
\\
{\hspace{0.5em}Full model}
& 23.2M
& 12.4G
& \textbf{82.45 \footnotesize {($+$2.62)}}
\\
\midrule
\multicolumn{4}{l}{\emph{Non-recursive + Soft Mixture}} \\
$D_{\mathrm{eff}}{=}22$
& 39.8M
& 8.4G
& 80.53 \footnotesize {($+$0.70)}
\\
$D_{\mathrm{eff}}{=}32$
& 57.6M
& 12.2G
& 80.51 \footnotesize {($+$0.68)}
\\
\bottomrule
\end{tabular}
}
\end{table}

\begin{table}[t]
\centering
\small
\caption{
Comparison between hard selection and the proposed Soft Mixture under different recursion depths.
}
\label{tab:mixture-ablation}
\begin{tabular}{lccc}
\toprule
\multirow{2}{*}{Aggregation Strategy}
& \multicolumn{3}{c}{Top-1 (\%)} \\
\cmidrule(lr){2-4}
& $T{=}1$ & $T{=}2$ & $T{=}3$ \\
\midrule
Hard Selection
& 79.83
& 81.79
& 80.54
\\
Soft Mixture (Ours)
& 79.83
& \textbf{82.02}
& \textbf{82.45}
\\
\bottomrule
\end{tabular}
\end{table}

\paragraph{Hard Selection vs. Soft Mixture.}
We further compare soft mixture with a hard selection variant in \cref{tab:mixture-ablation}. Hard selection follows the hard depth selection used in MoR: for each token, it selects only the recursion-step output with the highest routing score, instead of softly aggregating outputs from all recursion steps. When $T{=}1$, no recursion is performed and both variants reduce to the DeiT-S baseline. At $T{=}2$, hard selection improves over both the baseline and naive recursion, indicating that selecting among recursive-step outputs can be useful at shallow recursion depth. However, when $T$ is increased to 3, hard selection drops to 80.54\%, while Soft Mixture continues to improve and reaches 82.45\%. This result suggests that selecting only one recursion-step output is insufficient for reliably benefiting from deeper recursive computation. In contrast, Soft Mixture allows each token to combine information from multiple recursion steps, enabling intermediate recursive representations to contribute directly to the final output.

% \begin{table*}[t]
% \centering
% \caption{Effect of recursion depth on classification, detection, instance segmentation, and semantic segmentation performance.}
% \label{tab:recursion_depth}

% \begin{tabular}{lcccccc}
% \toprule

% Recursion ($T$) &
% Eff. depth &
% ImageNet &
% ImageNet (Distill) &
% COCO Det. &
% COCO Inst. &
% ADE20K Seg. \\

% \midrule

% Baseline DeiT-S
% & 12
% & 79.83
% & 81.18
% & 42.80
% & 38.38
% & 43.80
% \\

% $T{=}2$
% & 22
% & 82.02
% & 82.67
% & 44.63
% & 40.05
% & 45.81
% \\

% $T{=}3$
% & 32
% & 82.45
% & 83.36
% & 45.54
% & 40.78
% & 47.44
% \\

% $T{=}4$
% & 42
% & \textbf{82.48}
% & \textbf{83.57}
% & \textbf{46.25}
% & \textbf{41.32}
% & \textbf{47.96}
% \\

% \bottomrule
% \end{tabular}
% \end{table*}

\begin{figure}
    \centering
    \includegraphics[width=1\linewidth]{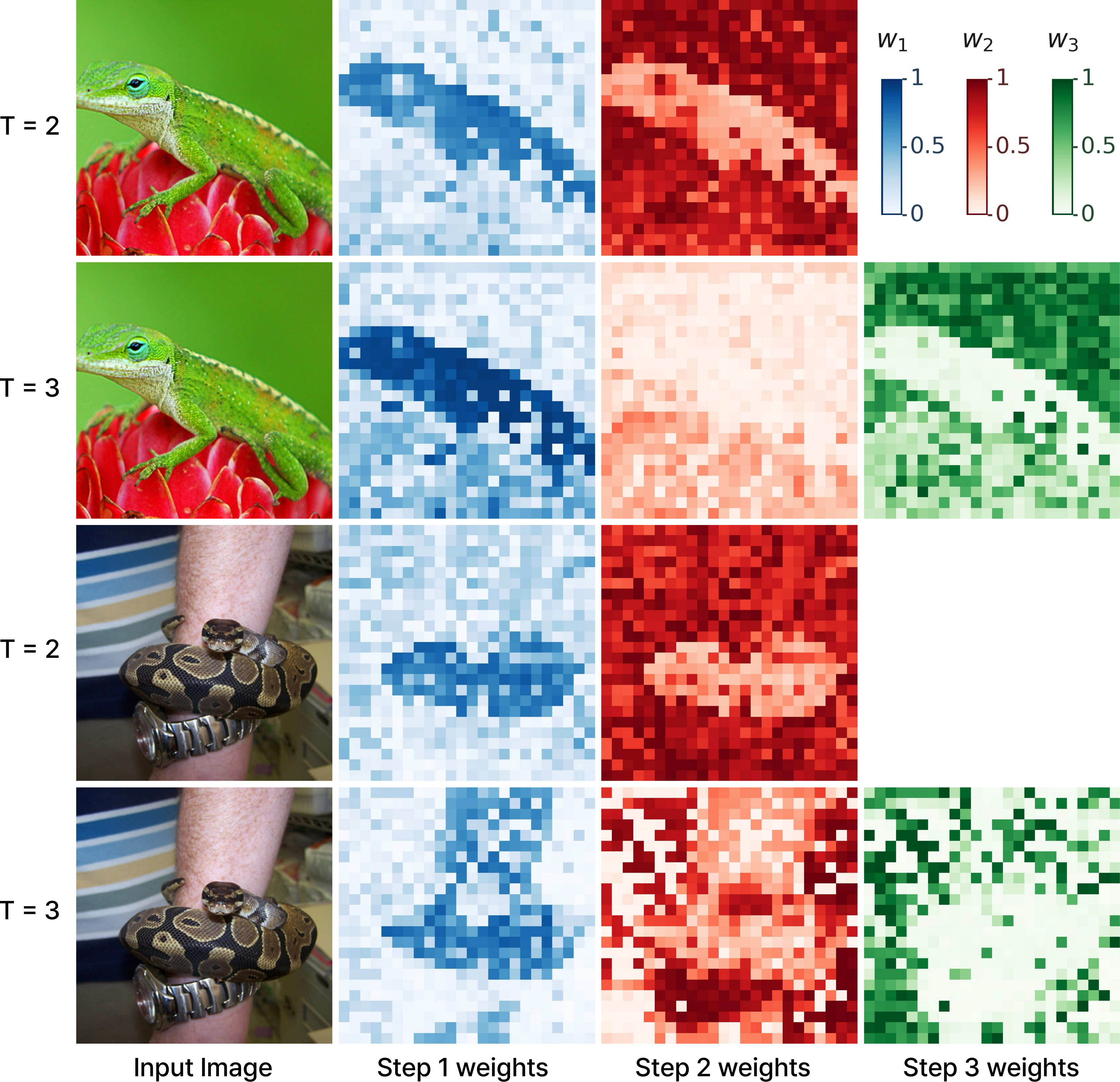}
    \caption{Visualization of learned token-wise mixture weights from the final recursive unit of SR-ViT-S. Blue, red, and green maps correspond to recursion steps 1, 2, and 3, respectively, with darker colors indicating larger weights. More examples on higher-resolution images are in Appendix \cref{sec:supp_mixture_vis}.}
    \label{fig:mixture-weights-vis}
\end{figure}

\paragraph{Visualization of Mixture Weights.}
To better understand how Soft Mixture uses recursive outputs, we visualize the learned token-wise mixture weights from the final recursive unit in \cref{fig:mixture-weights-vis}. Although no explicit visual or 2D guidance is imposed on the mixture weights, they form spatially coherent patterns that align with meaningful image regions. In the $T{=}2$ examples, the two recursion steps show a relatively discrete separation between target-object and background regions. In the $T{=}3$ examples, the patterns become more diverse: some regions are dominated by a single step, while other regions appear to combine multiple recursion steps. For example, in the $T{=}3$ row of the lizard example, Step~1 emphasizes the target object, whereas Step~2 and Step~3 emphasize different background regions. This suggests that, as recursion depth increases, multiple recursive outputs can contribute jointly to the final representation. Such behavior is difficult to capture with hard selection, which keeps only the maximum-weight step for each token. This qualitative behavior is consistent with \cref{tab:mixture-ablation}, where Soft Mixture continues to improve at $T{=}3$, while hard selection does not.

% \begin{table}[t]
% \centering
% \caption{Parameter-efficient variants of the proposed recursive architecture.
% Our reduced models maintain competitive performance while significantly reducing model size.}
% \label{tab:param_efficiency}

% \begin{tabular}{lcccc}
% \toprule

% Model &
% Depth &
% Params (M) &
% Top-1 (\%) &
% Param Reduction \\

% \midrule

% DeiT-S
% & 12
% & 22.1
% & 79.8
% & --
% \\

% Ours-S(2/3)-R2
% & 14
% & 15.4
% & 79.89
% & -30.3\%
% \\

% Ours-S(3/4)-R2
% & 16
% & 17.2
% & 80.82
% & -22.2\%
% \\

% Ours-S-R2
% & 22
% & 22.7
% & \textbf{82.02}
% & +2.7\%
% \\

% \bottomrule
% \end{tabular}
% \end{table}

%% file: sec/5_conclusion.tex
\section{Conclusion}
\label{sec:5}
% We presented Soft Mixture-of-Recursions (SoftMoR) and the Soft Recursive Vision Transformer (SR-ViT), which explore recursion as a parameter-efficient scaling axis for Vision Transformers. Instead of using only the final recursive output or assigning each token a single depth, SoftMoR learns token-wise mixture weights and softly combines representations from all recursion steps, allowing intermediate features from different effective depths to contribute directly to the final representation. SR-ViT reuses standard ViT layers across recursion steps, increasing effective computational depth with only marginal growth in the number of unique parameters. Experiments on ImageNet-1K, COCO, and ADE20K show favorable scaling behavior with recursion depth across classification and dense prediction tasks. Ablation studies further show that recursion depth alone is insufficient for effective scaling, while soft aggregation helps exploit the intermediate representations produced along the recursive path. Overall, our results support recursive deepening as a useful, parameter-efficient scaling axis for Vision Transformers.

We presented Soft Mixture-of-Recursions (SoftMoR) and the Soft Recursive Vision Transformer (SR-ViT) to investigate recursion as a scaling axis for Vision Transformers. Rather than stacking additional independently parameterized layers, SR-ViT increases effective computational depth by repeatedly applying shared Transformer blocks. However, our findings show that deeper recursion alone does not reliably yield stronger representations. SoftMoR addresses this limitation by learning to combine intermediate outputs across recursion steps, rather than retaining only the final or a selected state. Experiments on ImageNet-1K, COCO, and ADE20K demonstrate favorable scaling across classification and dense prediction with modest parameter growth, while ablations highlight the importance of effectively utilizing recursive representations. Although evaluated on vision tasks, SoftMoR does not rely on vision-specific inductive biases, leaving open the possibility of applying it to recursive Transformers in other domains. Overall, our findings suggest that recursion is more than a mechanism for parameter sharing: it offers a way to scale computation without proportionally scaling parameters, provided that the representations along the way are effectively utilized.

%% file: sec/supp.tex
\section{Implementation Details of SoftMoR}
\label{sec:supp_impl}

\cref{alg:softmor} summarizes the implementation of a SoftMoR unit. A SoftMoR unit consists of $K$ shared Transformer blocks that are recursively applied for $T$ recursion steps. The Soft Router predicts token-wise mixture weights over recursion steps from the input token features, and the outputs from all recursion steps are accumulated by a token-wise weighted sum. A lightweight linear Step Transform is applied between consecutive recursion steps to encourage step-specific variation.

\begin{algorithm}[H]
\caption{SoftMoR Unit}
\label{alg:softmor}
\begin{algorithmic}[1]
\State \textbf{Input:} token features $x \in \mathbb{R}^{B \times N \times D}$
\State \textbf{Output:} token features $y \in \mathbb{R}^{B \times N \times D}$
\State \textbf{Block configuration:} unit size $K$, recursion depth $T$
\vspace{0.3em}
\State $\alpha_1,\ldots,\alpha_T \leftarrow \texttt{SoftRouter}(x)$
\State $h \leftarrow x$
\State $y \leftarrow 0$
\For{$t = 1,\ldots,T$}
    \If{$t > 1$}
        \State $h \leftarrow h + \texttt{StepTransform}_{t}(h)$
    \EndIf
    \State $z \leftarrow h$
    \For{$k = 1,\ldots,K$}
        \State $z \leftarrow \texttt{TransformerBlock}_{k}(z)$
    \EndFor
    \State $y \leftarrow y + \alpha_t * z$
    \State $h \leftarrow z$
\EndFor
\State \textbf{return} $y$
\end{algorithmic}
\end{algorithm}

%-------------------------------------------------------------------------
\begin{figure*}[t]
    \centering
    \includegraphics[width=1\linewidth]{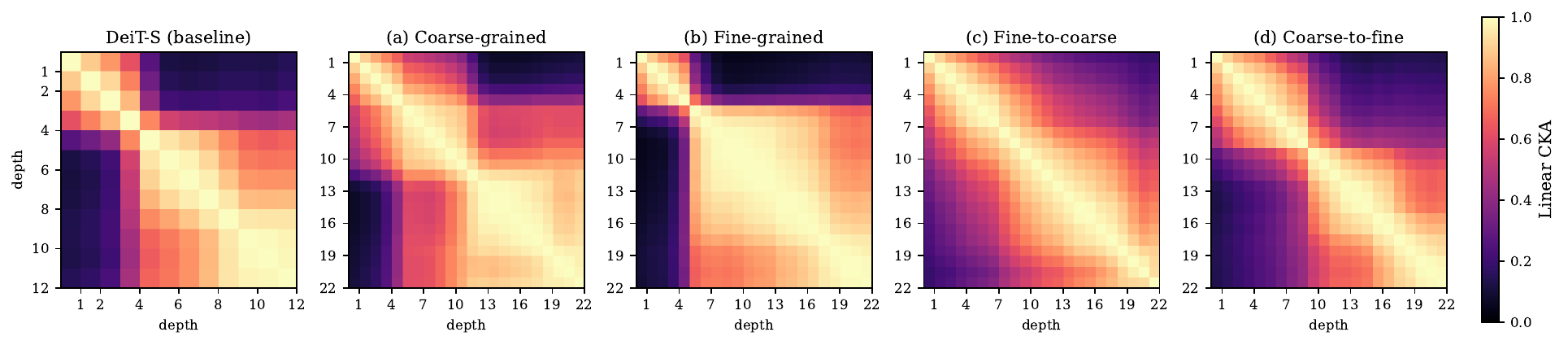}
    \vspace{-20pt}
    \caption{Token-level linear CKA across depth for DeiT-S and recursive ViT variants with different recursive unit organizations. DeiT-S contains 12 standard Transformer blocks, while the recursive variants are unrolled at recursion depth $T{=}2$, resulting in 22 effective depths. Brighter colors indicate higher representation similarity.}
    \label{fig:cka-analysis}
    \vspace{-10pt}
\end{figure*}

\begin{figure}
    \centering
    \includegraphics[width=1\linewidth]{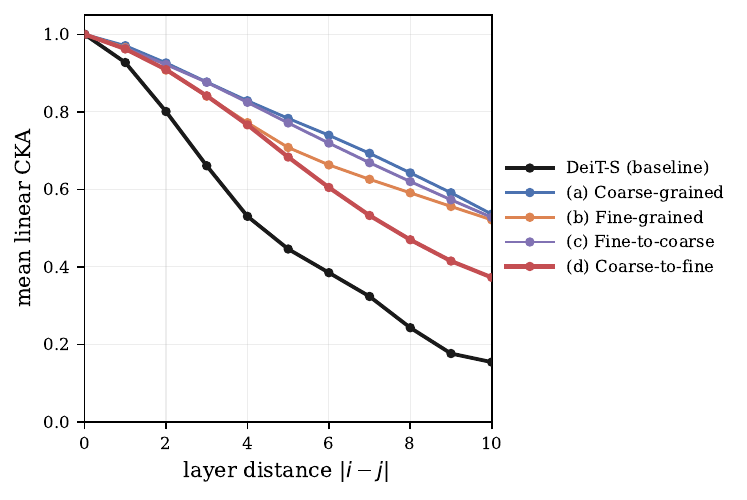}
    \caption{Mean token-level linear CKA according to layer distance $|i-j|$. Each point represents the average CKA value of all layer pairs separated by the same distance. A faster decrease indicates that representations change more across depth.}
    \label{fig:mean-cka}
\end{figure}

\section{Additional Analysis of Recursive Unit Organization}
\label{sec:supp_org}

The main paper compares several recursive unit organizations and shows that the coarse-to-fine strategy achieves the best ImageNet-1K~\cite{deng2009imagenet} accuracy among the evaluated variants. To further examine this behavior, we analyze token-level representation similarity using linear centered kernel alignment (CKA)~\cite{kornblith2019similarity}. CKA measures the similarity between neural representations by comparing the similarity structures induced by their activations, making it suitable for examining how representations evolve across layers or computational steps. For DeiT-S~\cite{touvron2021training}, CKA is computed across its 12 standard Transformer blocks. For the recursive variants, we unroll the computation path at recursion depth $T{=}2$, resulting in 22 effective depths.

As shown in \cref{fig:cka-analysis}, increasing effective depth through recursion leads to different representation dynamics depending on how the recursive units are organized. The coarse-grained organization exhibits block-like similarity regions, where the transition to the next recursion step appears relatively abrupt and less smoothly connected to the preceding step. This suggests that large recursive units may produce representations that behave more like separated recursive stages rather than a continuous depth-wise progression. The fine-grained organization shows broad high-similarity regions after the early depths, indicating that repeatedly applying very small recursive units can produce redundant representations. The fine-to-coarse organization also exhibits relatively strong similarity across wide depth ranges, together with less clear progressive transitions, suggesting that the increased effective depth does not necessarily translate into more diverse or gradually refined representations.

In contrast, the coarse-to-fine organization shows a more gradual diagonal structure that is visually closer to the non-recursive DeiT-S baseline, while extending the effective computational depth. This pattern suggests a more progressive evolution of token representations across depth, rather than overly redundant or abrupt changes. This qualitative pattern is consistent with the performance trend in \cref{tab:recursive-unit-acc}, where the coarse-to-fine organization achieves the highest accuracy.

We further summarize this trend in \cref{fig:mean-cka} by plotting the mean CKA value according to the layer distance $|i-j|$. Each point represents the average similarity between layer pairs separated by the same distance. If the CKA value remains high even for distant layers, the representations are still similar across depth, indicating that the additional computation may be redundant. In contrast, a faster decrease means that representations become more different as depth increases. Among the recursive variants, the coarse-to-fine organization shows the fastest decrease and follows the non-recursive DeiT-S trend most closely. This supports the observation that coarse-to-fine recursion leads to more progressive representation changes than the other recursive organizations.

We view this CKA analysis as qualitative diagnostic evidence that recursive unit organization affects how token-level representational similarity evolves across effective depth, which may influence whether additional recursive computation produces diverse and progressively refined visual representations.

\begin{table}
\centering
\caption{ImageNet-1K top-1 accuracy for different recursive unit organizations at $T{=}2$.}
\label{tab:recursive-unit-acc}
\begin{tabular}{lc}
\toprule
Model & Top-1 (\%) \\
\midrule
DeiT-S (baseline) & 79.83 \\
(a) Coarse-grained & 80.48 \\
(b) Fine-grained & 80.33 \\
(c) Fine-to-coarse & 78.89 \\
(d) Coarse-to-fine & \textbf{81.12} \\
\bottomrule
\end{tabular}
\vspace{-10pt}
\end{table}

\section{Architectural Specifications}
\label{sec:supp_specs}

We summarize the SR-ViT variants in \cref{tab:model-specs-supp}. All models keep the first and last Transformer blocks non-recursive and apply recursion to the remaining blocks with a coarse-to-fine unit layout. SR-ViT-T uses eight unique blocks arranged as $\mathrm{S}\text{-}R_{3}\text{-}R_{2}\text{-}R_{1}\text{-}\mathrm{S}$, while SR-ViT-S and SR-ViT-B use twelve unique blocks arranged as $\mathrm{S}\text{-}R_{4}\text{-}R_{3}\text{-}R_{2}\text{-}R_{1}\text{-}\mathrm{S}$. The MLP hidden size is $4\times$ the embedding dimension, and the attention head dimension is 64. Increasing $T$ raises $D_{\mathrm{eff}}$ and FLOPs, while the number of unique Transformer blocks stays fixed. Parameter counts grow only slightly with $T$ because of the lightweight Step Transform and Soft Mixture modules.

\begin{table*}[t]
\centering
\caption{Architectural specifications of SR-ViT variants. $S$ denotes a standard non-recursive Transformer block, and $R_K$ denotes a recursive unit containing $K$ sequential Transformer blocks.
Blocks denotes the number of unique Transformer blocks with independent parameters, while $D_{\mathrm{eff}}$ denotes the effective computational depth.
The MLP hidden size is $4\times$ the embedding dimension. FLOPs are measured at $224\times224$ with patch size $16\times16$.}
\label{tab:model-specs-supp}
\begin{tabular}{lcccccccc}
\toprule
Model & Recursion ($T$) & Structure & Blocks & $D_{\mathrm{eff}}$ & Embed dim & Heads & Params & FLOPs \\
\midrule
\multirow{2}{*}{SR-ViT-T}
& 2
& \multirow{2}{*}{$\mathrm{S}\text{-}R_{3}\text{-}R_{2}\text{-}R_{1}\text{-}\mathrm{S}$}
& \multirow{2}{*}{8}
& 14
& \multirow{2}{*}{384}
& \multirow{2}{*}{6}
& 15.4M
& 5.4G
\\
& 3
&
&
& 20
&
&
& 15.8M
& 7.8G
\\
\midrule
\multirow{3}{*}{SR-ViT-S}
& 2
& \multirow{3}{*}{$\mathrm{S}\text{-}R_{4}\text{-}R_{3}\text{-}R_{2}\text{-}R_{1}\text{-}\mathrm{S}$}
& \multirow{3}{*}{12}
& 22
& \multirow{3}{*}{384}
& \multirow{3}{*}{6}
& 22.7M
& 8.5G
\\
& 3
&
&
& 32
&
&
& 23.2M
& 12.4G
\\
& 4
&
&
& 42
&
&
& 23.8M
& 16.3G
\\
\midrule
\multirow{2}{*}{SR-ViT-B}
& 2
& \multirow{2}{*}{$\mathrm{S}\text{-}R_{4}\text{-}R_{3}\text{-}R_{2}\text{-}R_{1}\text{-}\mathrm{S}$}
& \multirow{2}{*}{12}
& 22
& \multirow{2}{*}{512}
& \multirow{2}{*}{8}
& 39.9M
& 14.8G
\\
& 3
&
&
& 32
&
&
& 41.0M
& 21.6G
\\
\bottomrule
\end{tabular}
\vspace{-5pt}
\end{table*}

%------------------------------------------------------------------------
\section{Detailed Experimental Settings}
\label{sec:supp_exp_details}

\begin{table}[t]
\centering
\caption{ImageNet-1K training recipe used for SR-ViT.}
\label{tab:imagenet_recipe}
\begin{tabular}{ll}
\toprule
Setting & Value \\
\midrule
Epochs & 300 \\
Batch size & 1024 \\
Optimizer & AdamW \\
Learning rate & 0.001 \\
LR schedule & Cosine \\
Warmup epochs & 5 / 10 / 20 for $T{=}2/3/4$ \\
Weight decay & 0.05 \\
Augmentation & RandAugment \\
Repeated augmentation & 3 repeats \\
Random erasing & 0.25 \\
Mixup & 0.8 \\
CutMix & 1.0 \\
Label smoothing & 0.1 \\
Eval crop ratio & 0.875 \\
\bottomrule
\end{tabular}
\vspace{-10pt}
\end{table}

\subsection{ImageNet-1K Classification}

We train SR-ViT on ImageNet-1K following the standard DeiT training recipe~\cite{touvron2021training}. All models are trained from scratch for 300 epochs using AdamW with a global batch size of 1024 and a learning rate of 0.001. We use the same augmentation and regularization settings across model variants, including RandAugment~\cite{cubuk2020randaugment}, repeated augmentation~\cite{hoffer2020augment}, random erasing~\cite{zhong2020random}, Mixup~\cite{zhang2018mixup}, CutMix~\cite{yun2019cutmix}, and label smoothing~\cite{muller2019does}. To improve training stability for deeper recursive models, we use longer warmup schedules as recursion depth increases. For 384-resolution evaluation, we fine-tune the 224-resolution pretrained models for 30 epochs with a global batch size of 512 and a learning rate of $1 \times 10^{-5}$, using an evaluation crop ratio of 1.0. For DeiT-style distillation, we follow the original DeiT setting and use a RegNetY-16GF~\cite{radosavovic2020designing} teacher with hard distillation. The ImageNet-1K training recipe is summarized in \cref{tab:imagenet_recipe}.

We additionally compare SR-ViT with representative ImageNet-1K backbones at 224 resolution in \cref{tab:imagenet-perfornamce}, including recent recursive Vision Transformers. SR-ViT remains competitive among plain Transformers while using substantially fewer parameters than DeiT-B.

\subsection{COCO Object Detection and Instance Segmentation}

The main paper reports the COCO object detection and instance segmentation results. Here, we provide the detailed fine-tuning setup used for all COCO experiments.

We evaluate object detection and instance segmentation on COCO 2017~\cite{lin2014microsoft} using Mask R-CNN~\cite{he2017mask} with a ViTDet-style training setup~\cite{li2022exploring}. ImageNet-1K pretrained backbones are integrated with a Simple Feature Pyramid constructed from the final backbone feature map. For both DeiT and SR-ViT backbones, the final patch-token features are reshaped into a single spatial feature map with stride 16, after removing class tokens. The Simple Feature Pyramid then constructs multi-scale features using output channels of 256 and scale factors of $4.0$, $2.0$, $1.0$, and $0.5$.

All models are fine-tuned with Large Scale Jittering (LSJ)~\cite{ghiasi2021simple} augmentation using $1024 \times 1024$ training crops. We use the standard $3\times$ schedule with 270k iterations, a global batch size of 16, AdamW optimizer, and an initial learning rate of $1 \times 10^{-4}$. The learning rate follows a cosine decay schedule with a 1000-iteration warmup. We also use layer-wise learning-rate decay with a decay rate of 0.7. Mixed-precision training is enabled for all experiments.

To ensure a fair comparison, all detection and instance segmentation settings are kept identical across backbones. The only difference between models is the backbone architecture.

\subsection{ADE20K Semantic Segmentation}

The main paper reports the ADE20K semantic segmentation results. Here, we provide the detailed fine-tuning setup used for all ADE20K experiments.

We evaluate semantic segmentation on ADE20K~\cite{zhou2017scene} using UPerNet~\cite{xiao2018unified} with MMSegmentation~\cite{mmseg2020}. All ImageNet-1K pretrained backbones are integrated with a Multi-Level Neck (MLN) to construct multi-scale features. For DeiT backbones, we extract intermediate features from Transformer layers at depths $L/4$, $2L/4$, $3L/4$, and $L$, corresponding to layers 3, 6, 9, and 12 for 12-layer DeiT models. For SR-ViT, we use the outputs of the four recursive units $(R_4, R_3, R_2, R_1)$ as hierarchical feature representations. The MLN uses scale factors of $4$, $2$, $1$, and $0.5$.

All models are trained under the same 160K iteration schedule with AdamW, an initial learning rate of $6 \times 10^{-5}$, weight decay of 0.01, and a global batch size of 16. The learning rate follows a polynomial decay schedule after a 1500-iteration linear warmup. We use standard ADE20K data augmentation, including random resizing, random cropping with a crop size of $512 \times 512$, random horizontal flipping, and photometric distortion.

For evaluation, we report both single-scale and multi-scale mIoU. Multi-scale evaluation uses scale ratios of ${0.5, 0.75, 1.0, 1.25, 1.5, 1.75}$ with horizontal flipping. To ensure a fair comparison, all segmentation settings are kept identical across backbones, with the backbone architecture being the only variable.

\section{Efficiency Analysis of Recursive Deepening}
\label{sec:supp_efficiency}

We examine how the computational cost of SR-ViT changes as recursion depth $T$ increases. \cref{tab:supp-efficiency} compares SR-ViT-S at $T{=}2/3/4$ with independent-layer ViT-S models that apply the same number of Transformer blocks without weight sharing. This isolates recursive deepening from simply stacking additional layers.

\begin{table*}[t]
\centering
\small
\setlength{\tabcolsep}{5pt}
\caption{Throughput and GPU memory at $224\times224$ on an RTX~4090. Memory is reported in MiB. Dim is the embedding dimension. ImageNet Top-1 accuracy is included as a reference. ViT-S-D22/D32/D42 are independent-layer ViTs with the same unrolled depth as SR-ViT-S at $T{=}2/3/4$.}
\label{tab:supp-efficiency}
\begin{tabular}{lrrrrrrrc}
\toprule
Model & Dim & Params & $D_{\mathrm{eff}}$ & FLOPs & Throughput (img/s) & Inf.\ mem. & Train mem. & Top-1 (\%) \\
\midrule
\multicolumn{9}{l}{\textit{Non-recursive baseline}} \\
DeiT-S & 384 & 22.1M & 12 & 4.6G & 2614 & 352 & 2140 & 79.83 \\
\midrule
\multicolumn{9}{l}{\textit{$D_{\mathrm{eff}}{=}22$}} \\
ViT-S-D22 & 384 & 39.8M & 22 & 8.4G & 1433 & 420 & 3826 & 79.73 \\
\rowcolor{blue!5}
SR-ViT-S ($T{=}2$) & 384 & 22.7M & 22 & 8.5G & 1370 & 430 & 3815 & 82.02 \\
\midrule
\multicolumn{9}{l}{\textit{$D_{\mathrm{eff}}{=}32$}} \\
ViT-S-D32 & 384 & 57.5M & 32 & 12.2G & 991 & 488 & 5513 & 78.60 \\
\rowcolor{blue!5}
SR-ViT-S ($T{=}3$) & 384 & 23.2M & 32 & 12.4G & 943 & 452 & 5380 & 82.45 \\
\midrule
\multicolumn{9}{l}{\textit{$D_{\mathrm{eff}}{=}42$}} \\
ViT-S-D42 & 384 & 75.3M & 42 & 16.0G & 755 & 556 & 7199 & -- \\
\rowcolor{blue!5}
SR-ViT-S ($T{=}4$) & 384 & 23.8M & 42 & 16.3G & 718 & 455 & 6944 & 82.48 \\
\midrule
\multicolumn{9}{l}{\textit{Width-scaled models}} \\
DeiT-B & 768 & 86.6M & 12 & 17.5G & 881 & 820 & 4730 & 81.80 \\
SR-ViT-B ($T{=}2$) & 512 & 39.9M & 22 & 14.8G & 898 & 598 & 5211 & 82.76 \\
SR-ViT-B ($T{=}3$) & 512 & 41.0M & 32 & 21.6G & 620 & 627 & 7311 & 82.19 \\
\bottomrule
\end{tabular}
\end{table*}

As shown in \cref{tab:supp-efficiency}, increasing $T$ raises $D_{\mathrm{eff}}$ and FLOPs, and throughput follows this additional computational depth. At $224\times224$, SR-ViT-S goes from 1370~img/s at $T{=}2$ to 943 and 718~img/s at $T{=}3$ and $T{=}4$, closely matching the independent-layer models at the same $D_{\mathrm{eff}}$. The remaining 4--5\% difference is consistent with the additional computation of Step Transform and Soft Mixture. Recursive deepening therefore has the same compute--throughput behavior as a deeper Transformer stack, while keeping the unique parameter count nearly unchanged: SR-ViT-S uses 22.7M parameters at $T{=}2$ and only 23.8M at $T{=}4$. Additional recursive depth increases computation without expanding the set of unique Transformer blocks, providing a parameter-efficient way to scale effective depth. Top-1 accuracy is included as a reference; independently stacked models at the same $D_{\mathrm{eff}}$ do not benefit from the extra depth, consistent with prior observations that naively stacking ViT layers is not an effective scaling axis~\cite{zhou2021deepvit}.

Training memory also tracks $D_{\mathrm{eff}}$ and remains comparable to the independent-layer models, with a small but consistent advantage at each depth. Inference memory benefits more clearly from recursion. From $T{=}2$ to $T{=}4$, SR-ViT-S increases only from 430 to 455~MiB, whereas stacking independent layers grows inference memory from 420 to 556~MiB due to the increased parameters. Recursive deepening thus scales computational depth with little additional inference-memory cost relative to adding new layers.

We further compare these depth-scaled models with DeiT-B. At similar FLOPs, DeiT-B attains higher throughput than ViT-S-D42 (881 vs.\ 755~img/s) and uses less training memory (4730 vs.\ 7199~MiB). SR-ViT-B at $T{=}2$ uses fewer parameters and FLOPs than DeiT-B, while reaching a similar throughput (898 vs.\ 881~img/s) with lower inference memory (598 vs.\ 820~MiB). Compared with adding more layers to increase depth, recursive deepening follows a similar compute--throughput trade-off, while inference memory stays nearly unchanged even as depth grows; keeping a small parameter count is the most attractive benefit.

\begin{table}[t]
\centering
\caption{ImageNet-1K top-1 accuracy of aggregation strategies and Step Transform on SR-ViT-S. $T{=}1$ corresponds to DeiT-S.}
\label{tab:supp-aggregation}
\resizebox{\linewidth}{!}{
\begin{tabular}{lcccc}
\toprule
\multirow{2}{*}{Aggregation} &
\multirow{2}{*}{Step Transform} &
\multicolumn{3}{c}{Top-1 (\%)} \\
\cmidrule(lr){3-5}
& & $T{=}1$ & $T{=}2$ & $T{=}3$ \\
\midrule
\multirow{2}{*}{Uniform average}
& \xmark & 79.83 & 80.56 & 81.55 \\
& \checkmark & 79.83 & 79.32 & 77.99 \\
\midrule
\multirow{2}{*}{Hard selection}
& \xmark & 79.83 & 81.64 & 80.96 \\
& \checkmark & 79.83 & 81.79 & 80.54 \\
\midrule
\multirow{2}{*}{Soft Mixture}
& \xmark & 79.83 & 81.59 & 81.74 \\
& \checkmark & 79.83 & \textbf{82.02} & \textbf{82.45} \\
\bottomrule
\end{tabular}
}
\vspace{-5pt}
\end{table}

\section{Aggregation Strategies and Step Transform}
\label{sec:supp_aggregation}

MoR-style hard selection assigns each token a single recursion depth and uses only that step's output. The remaining recursive-step representations do not contribute to the final output. SoftMoR instead learns token-wise mixture weights over all recursion steps, so that each token can use every depth in different proportions. \cref{tab:supp-aggregation} compares this soft mixture with hard selection and with a uniform average of all step outputs, with and without Step Transform. When $T{=}1$, no recursion is applied and all variants reduce to DeiT-S (79.83\%).

\begin{table}[t]
\centering
% \small
\caption{ImageNet-1K comparison at $224$ resolution with representative vision backbones and recent recursive approaches. Models marked with $\dagger$ use DeiT-style hard distillation.}
\label{tab:imagenet-perfornamce}
\begin{tabular}{lrrc}
\toprule
Model & Params & FLOPs & Top-1 (\%) \\
\midrule
\multicolumn{4}{l}{\textit{Conv-based models}} \\
ConvNeXt-T~\cite{liu2022convnet} & 29M & 4.5G & 82.1 \\
ConvNeXt-S~\cite{liu2022convnet} & 50M & 8.7G & 83.1 \\
ConvNeXt-B~\cite{liu2022convnet} & 89M & 15.4G & 83.8 \\
\midrule
\multicolumn{4}{l}{\textit{SSM-based models}} \\
Vim-T~\cite{zhu2024vision} & 7M & 1.8G & 76.1 \\
Vim-S~\cite{zhu2024vision} & 26M & 5.9G & 80.3 \\
Vim-B~\cite{zhu2024vision} & 98M & 20.9G & 81.9 \\
\midrule
\multicolumn{4}{l}{\textit{Hierarchical Transformers}} \\
Swin-T~\cite{liu2021swin} & 28M & 4.5G & 81.3 \\
Swin-S~\cite{liu2021swin} & 50M & 8.7G & 83.0 \\
Swin-B~\cite{liu2021swin} & 88M & 15.4G & 83.5 \\
\midrule
\multicolumn{4}{l}{\textit{Plain Vision Transformers}} \\
DeiT-Ti~\cite{touvron2021training} & 6M & 1.3G & 72.2 \\
DeiT-S~\cite{touvron2021training} & 22M & 4.6G & 79.8 \\
DeiT-B~\cite{touvron2021training} & 87M & 17.5G & 81.8 \\
DeiT-Ti$^\dagger$~\cite{touvron2021training} & 6M & 1.3G & 74.5 \\
DeiT-S$^\dagger$~\cite{touvron2021training} & 22M & 4.6G & 81.2 \\
DeiT-B$^\dagger$~\cite{touvron2021training} & 87M & 17.5G & 83.3 \\
\midrule
\multicolumn{4}{l}{\textit{Recursive Vision Transformers}} \\
Vision-MoR-XL~\cite{he2026visionmor} & 14M & 2.0G & 80.4 \\
Edge-RecViT-S~\cite{li2026edgerecvit} & 6M & 3.2G & 80.3 \\
Edge-RecViT-B~\cite{li2026edgerecvit} & 23M & 12.2G & 82.0 \\
\midrule
\rowcolor{blue!5}
\multicolumn{4}{l}{\textit{Ours}} \\
\rowcolor{blue!5}
SR-ViT-T ($T{=}2$) & 15M & 5.4G & 79.9 \\
\rowcolor{blue!5}
SR-ViT-T ($T{=}3$) & 16M & 7.8G & 80.9 \\
\rowcolor{blue!5}
SR-ViT-S ($T{=}2$) & 23M & 8.5G & 82.0 \\
\rowcolor{blue!5}
SR-ViT-S ($T{=}3$) & 23M & 12.4G & 82.5 \\
\rowcolor{blue!5}
SR-ViT-B ($T{=}2$) & 40M & 14.8G & 82.8 \\
\rowcolor{blue!5}
SR-ViT-T$^\dagger$ ($T{=}3$) & 16M & 7.8G & 81.9 \\
\rowcolor{blue!5}
SR-ViT-S$^\dagger$ ($T{=}2$) & 23M & 8.5G & 82.7 \\
\rowcolor{blue!5}
SR-ViT-S$^\dagger$ ($T{=}3$) & 23M & 12.4G & 83.4 \\
\rowcolor{blue!5}
SR-ViT-S$^\dagger$ ($T{=}4$) & 24M & 16.3G & 83.6 \\
\rowcolor{blue!5}
SR-ViT-B$^\dagger$ ($T{=}2$) & 40M & 14.8G & 83.8 \\
\rowcolor{blue!5}
SR-ViT-B$^\dagger$ ($T{=}3$) & 41M & 21.6G & \textbf{84.0} \\
\bottomrule
\end{tabular}
\end{table}

Hard selection improves over DeiT-S at $T{=}2$, reaching 81.64\% without Step Transform and 81.79\% with it. When $T$ is increased to 3, however, accuracy drops to 80.96\% and 80.54\%. Assigning each token a single recursion depth is therefore insufficient to benefit from additional recursive computation.

A natural alternative is to use every recursive-step output. Uniform average does so by taking the mean of all step representations as the final output. This prevents the $T{=}3$ drop of hard selection and reaches 81.55\%, showing that intermediate steps contain useful information. At $T{=}2$, however, uniform average (80.56\%) remains below both hard selection and Soft Mixture, so using all steps with equal weights is not by itself the most effective combination.

Soft Mixture also uses all recursive-step outputs, but with learned token-wise weights. It improves from $T{=}2$ to $T{=}3$ both without Step Transform (81.59\% to 81.74\%) and with it (82.02\% to 82.45\%), and is the strongest method at both depths. The benefit of additional recursion therefore comes from using intermediate representations in a learnable and flexible way, rather than from selecting a single depth or averaging all depths equally.

Step Transform applies a recursion-specific map before the next application of the shared Transformer blocks; the blocks themselves remain strictly shared. It is not a generic gain from extra parameters. Adding Step Transform degrades uniform average (79.32\% and 77.99\%) and does not prevent the $T{=}3$ drop of hard selection (81.79\% to 80.54\%). In contrast, it improves Soft Mixture at both depths (81.59\% to 82.02\%, 81.74\% to 82.45\%). Step Transform is therefore most effective together with a learnable token-wise mixture. The two components are complementary: Step Transform encourages variation across recursion steps, and Soft Mixture lets each token use those representations at different effective depths according to learned relevance.

%------------------------------------------------------------------------
\section{More Visualization of Mixture Weights}
\label{sec:supp_mixture_vis}

\begin{figure*}[p]
    \centering
    \includegraphics[width=1\linewidth]{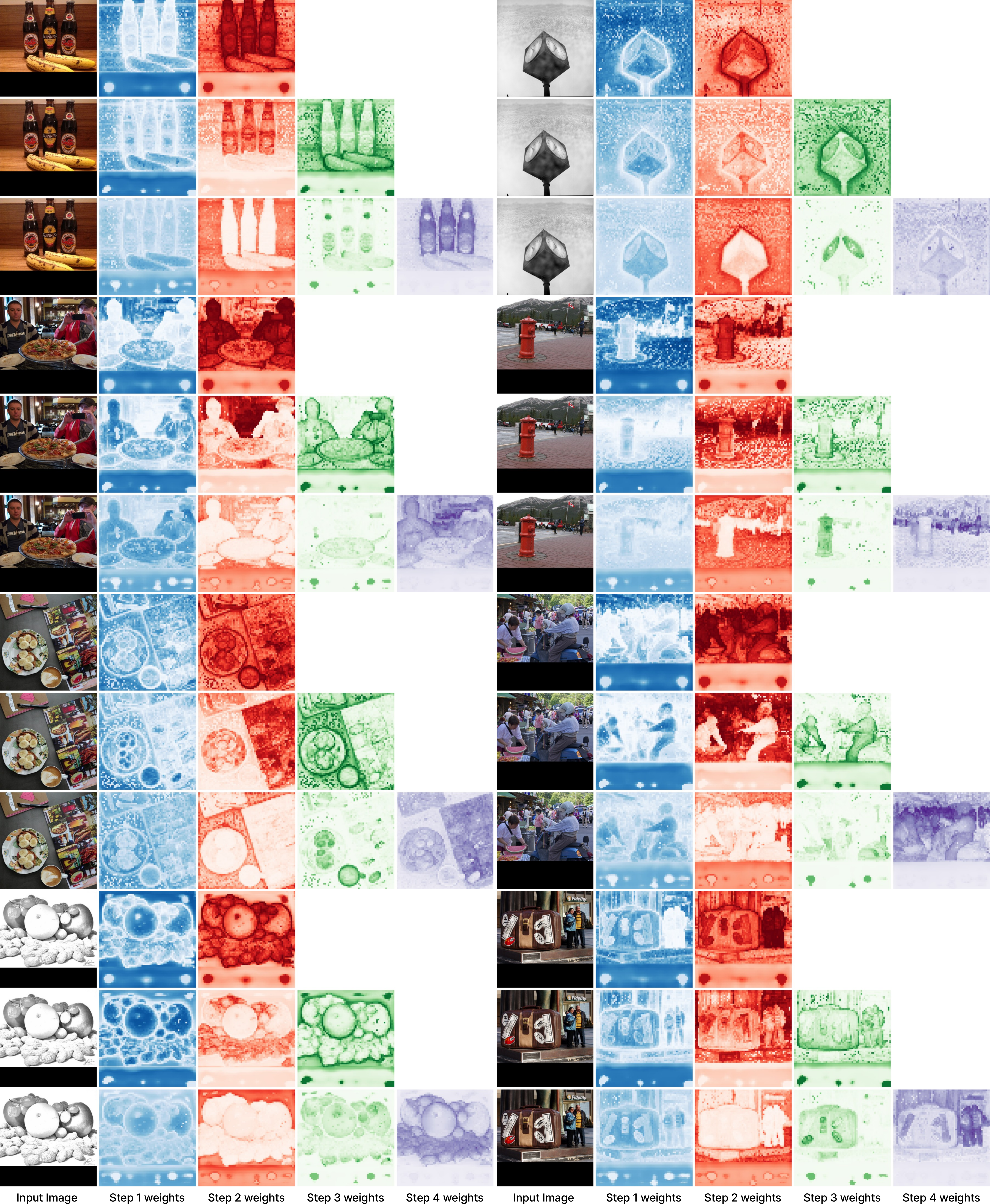}
    \caption{Additional visualizations of learned mixture weights on COCO. We visualize token-wise mixture weights from the final recursive unit of SR-ViT-S trained on COCO. The examples include models with recursion depths $T{=}2$, $T{=}3$, and $T{=}4$. Blue, red, green, and purple maps correspond to recursion steps 1, 2, 3, and 4, respectively, with darker colors indicating larger weights.}
    \vspace{-6pt}
    \label{fig:mixture-weights-coco}
\end{figure*}

\begin{figure*}[p]
    \centering
    \includegraphics[width=1\linewidth]{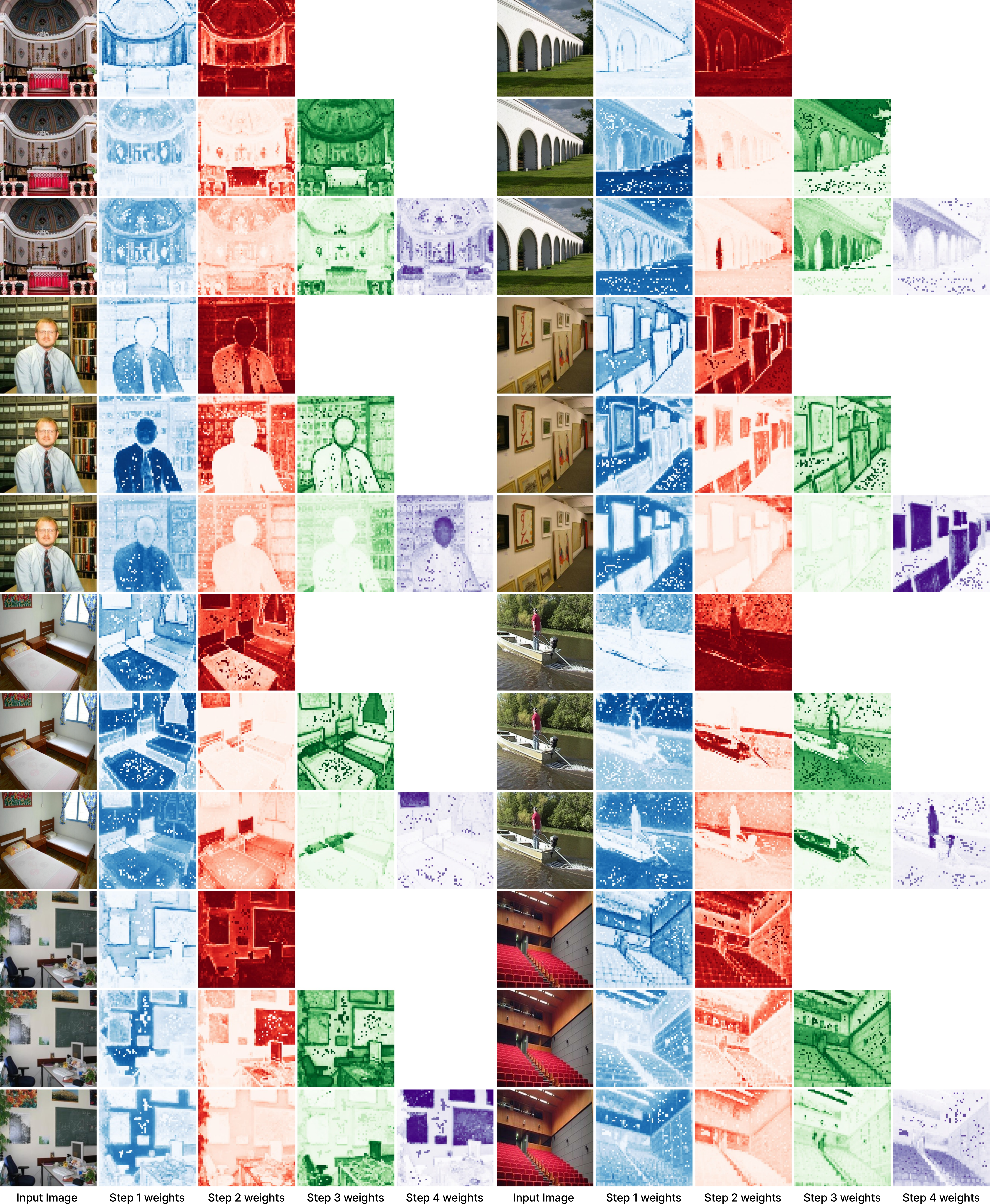}
    \caption{Additional visualizations of learned mixture weights on ADE20K. We visualize token-wise mixture weights from the final recursive unit of SR-ViT-S trained on ADE20K. The examples include models with recursion depths $T{=}2$, $T{=}3$, and $T{=}4$. Blue, red, green, and purple maps correspond to recursion steps 1, 2, 3, and 4, respectively, with darker colors indicating larger weights.}
    \vspace{-6pt}
    \label{fig:mixture-weights-ade}
\end{figure*}

We provide additional visualizations of the learned mixture weights in \cref{fig:mixture-weights-coco,fig:mixture-weights-ade}. Following the analysis in the main paper, we visualize the token-wise mixture weights from the final recursive unit. The examples are extracted from SR-ViT-S backbones trained on COCO and ADE20K, using recursion depths $T{=}2$, $T{=}3$, and $T{=}4$. Since the downstream models operate on high-resolution $1024 \times 1024$ inputs with a patch size of 16, the resulting mixture-weight maps provide a more detailed view of how SoftMoR assigns different recursion steps across spatial regions.

The visualizations show that the learned mixture weights form spatially coherent and semantically meaningful patterns, even though no explicit spatial supervision is applied to the mixture weights. Across both COCO and ADE20K examples, different recursion steps tend to emphasize different image regions, such as foreground objects, object boundaries, background regions, and visually complex areas. As the recursion depth increases, the mixture patterns become more diverse, suggesting that multiple recursive outputs contribute different types of information to the final representation.

These qualitative results support the role of SoftMoR as a learnable aggregation mechanism over recursive computation. Although the same Transformer blocks are reused across recursion steps, the combination of step transforms and soft mixture allows different recursion steps to take on distinct visual roles, leading to richer representations.

\clearpage

